\documentclass{article}
\usepackage{spconf,amsmath,graphicx}
\usepackage{bm}
\usepackage{amssymb}
\usepackage{cite}
\usepackage{subfig}
\usepackage[hidelinks]{hyperref}

\title{ON THE USE OF MODALITY-SPECIFIC LARGE-SCALE PRE-TRAINED ENCODERS\\FOR MULTIMODAL SENTIMENT ANALYSIS}
\name{\begin{tabular}{c}Atsushi Ando, Ryo Masumura, Akihiko Takashima, Satoshi Suzuki, Naoki Makishima,\\
		 Keita Suzuki, Takafumi Moriya, Takanori Ashihara, Hiroshi Sato\end{tabular}}
\address{NTT Corporation}

\begin{document}

%
\maketitle
%
\begin{abstract}
This paper investigates the effectiveness and implementation of modality-specific large-scale pre-trained encoders for multimodal sentiment analysis~(MSA).
Although the effectiveness of pre-trained encoders in various fields has been reported, conventional MSA methods employ them for only linguistic modality, and their application has not been investigated.
This paper compares the features yielded by large-scale pre-trained encoders with conventional heuristic features.
One each of the largest pre-trained encoders publicly available for each modality are used; CLIP-ViT, WavLM, and BERT for visual, acoustic, and linguistic modalities, respectively.
Experiments on two datasets reveal that methods with domain-specific pre-trained encoders attain better performance than those with conventional features in both unimodal and multimodal scenarios.
We also find it better to use the outputs of the intermediate layers of the encoders than those of the output layer.
The codes are available at \url{https://github.com/ando-hub/MSA_Pretrain}.
\end{abstract}


\section{Introduction}
\label{sec:intro}
Multimodal sentiment analysis~(MSA) is the technology to estimate the sentiment of a target speaker from multimodal information such as visual, acoustic, and linguistic modalities.
Since sentimental cues appear in various aspects such as facial expression, tone, and phrases, MSA performs better than alternatives using a single modality such as facial expression recognition~\cite{A_Survey_on_Human}, speech emotion recognition~\cite{Survey_on_Deep}, and sentiment analysis from text~\cite{A_Comprehensive_Survey}.

Most conventional studies have focused on modeling the interactions of multiple modalities and modality-specific information.
Some aim to model the interactions of short-term features of each modality~\cite{Words_Can_Shift, MulT}.
They can use local characteristics across modalities, e.g., facial expression changes or prosody during a particular word, for enhancing prediction performance.
Others extract sequence-level representations of the speaker's sentiment in the individual modalities, then estimate the sentiment level from all of the sequence-level representations~\cite{Tensor_Fusion_Network, Multimodal_Language_Analysis}.
These representations have been evaluated against each other to learn modality-invariant and modality-specific information, with the goal being to improve robustness against missing information of specific modalities such as facial occlusion~\cite{Learning_Factorized_Multimodal, Found_in_Translation}.
The studies employ heuristic features and\,/\,or the prediction results of the model such as head pose, gaze, and facial landmarks.

Recently, several MSA studies have utilized large-scale pre-trained encoders in addition to developing model structures~\cite{Integrating_Multimodal_Information, MAGplus, MISA, SelfMM}.
The pre-trained encoder is a part of the model trained in other tasks known as upstream tasks.
The advantage of the pre-trained encoder is that it enables the transfer of common knowledge of upstream tasks, which yields better cues for a target downstream task compared to training from scratch.
It has also been reported that a larger pre-trained encoder trained on a large amount of upstream data offers better performance in downstream tasks~\cite{GPT3, wav2vec2}.
Large-scale pre-training encoders have significantly enhanced various downstream tasks in visual, acoustic, and linguistic modalities~\cite{CLIP, wavlm, BERT}.
Though the introduction of the pre-trained encoder has improved MSA performance, the conventional studies suffer from two omissions.
First, they employ the large-scale pre-trained encoder only in linguistic modality, not in visual and acoustic modalities.
Second, how to apply the pre-trained encoders has not been investigated.
Some studies use the weighted sum results of the hidden states extracted from each layer of the pre-trained encoder for speech and speaker recognition~\cite{wavlm} since it is empirically known that different knowledge is extracted in the different layers in the pre-trained encoder~\cite{HuBERT, What_Does_BERT}.
However, the previous work employed only the output of the final layer of the pre-trained model, which may be less effective for MSA than the use of the hidden states of the intermediate layers.

This paper investigates the following two research questions: 
(i) Are the features based on the modality-specific pre-trained encoder more effective than the conventional features in multimodal, and even unimodal, scenarios?
(ii) How to apply the modality-specific pre-trained encoders?
Three large-scale pre-trained encoders that are currently available for each modality are examined; CLIP Vision Transformer~(ViT)~\cite{CLIP}, WavLM~\cite{wavlm}, and BERT~\cite{BERT} for visual, acoustic, and linguistic modalities, respectively.
This paper introduces a simple sequence-level cross-modal model, Unimodal Encoders and Gated Decoder~(UEGD), which enables comparison of multimodal and unimodal performances on the same model structure.
We evaluate three types of representations from the pre-trained encoders that have been used in other downstream tasks; the output, each of the intermediate outputs, and a weighted sum of the intermediate outputs.
Experiments on two public datasets, CMU-MOSI~\cite{mosi} and CMU-MOSEI\cite{CMUMOSEI}, reveal the answers to our research questions:
(i) Large-scale pre-trained encoders improve sentiment analysis performance in both unimodal and multimodal scenarios.
The UEGD model with pre-trained encoders achieves state-of-the-art performance in regression tasks on CMU-MOSEI.
It is also found that the pre-trained encoder is particularly effective in the acoustic modality.
(ii) Using one of the late middle intermediate layers of the pre-trained encoder yield better performance than the final layer output and the weighted sum of the intermediate layer outputs.

\section{Related Work}
\label{sec:related}
\subsection{Multimodal Sentiment Analysis}
\label{sec:related_msa}
The conventional MSA methods can be categorized into two approaches; early-fusion and late-fusion.

The early-fusion approach aims to capture interactive information from low-level features such as frame-level features in visual and acoustic modalities and word-level features in linguistic modality.
Multimodal Transformer~(MulT) employs multiple crossmodal transformers to capture bi-modal interactions~\cite{MulT}.
Multimodal Adaptation Gate~(MAG) focuses on integrating linguistic features with other modalities, visual and acoustic factors, by using a gating structure in addition to the cross-modal self-attention~\cite{Integrating_Multimodal_Information, MAGplus}.
The advantage of this approach lies in capturing local characteristics across modalities.
However, it can only be used when two or more modalities are available.

Late-fusion integrates utterance-level representations of modalities to predict sentiment.
Tensor Fusion Network~(TFN) explicitly models uni-, bi-, and tri-modal interactions as outer products of utterance-level embeddings~\cite{Tensor_Fusion_Network}.
Modality-Invariant and -Specific Representations~(MISA) extract modality-invariant\,/\,-specific utterance-level representations by introducing similarity and difference losses of representations~\cite{MISA}.
Self-Supervised Multi-task Multimodal sentiment analysis~(Self-MM) jointly learns multimodal and unimodal subtasks from utterance-level representations to supplement modality-specific information~\cite{SelfMM}.

The MSA model in this paper uses a late-fusion approach since it enables performance comparisons in unimodal and multimodal scenarios on the same model structure.

\subsection{Large-Scale Pre-Trained Encoders}
\label{sec:related_pt}
Large-scale pre-trained encoders have received significant attention in recent years.
In this framework, an upstream model is trained by a large amount of training data, then a small amount of labeled data is used to adapt downstream tasks in combination with a part of the upstream model.
The tasks that do not require human annotation, such as contrastive learning~\cite{simclr} or self-supervised learning~\cite{wav2vec2}, are used as the upstream task.
It has been reported that larger models trained by large amounts of upstream data show higher performance in many downstream tasks.
Pre-trained encoders comprising a stack of multiple transformer encoders~\cite{Attention_is_All} are often used.
This approach was first used with great success in natural language processing~\cite{BERT, xlnet} and is now widely used in computer vision~\cite{CLIP} and speech processing~\cite{HuBERT ,wavlm}.
It has been empirically reported that low-level features are extracted in the layers closer to the input, e.g., phrase-level information in linguistic encoder, and high-level features are obtained in the layers closer to the output like long-distance dependency information~\cite{What_Does_BERT, wavlm}.

This work employs the pre-trained models that are some of the largest and publicly available ones; CLIP-ViT~\cite{CLIP}, WavLM~\cite{wavlm}, and BERT~\cite{BERT} in visual, acoustic, and linguistic modalities, respectively.
CLIP-ViT extracts an image embedding from a single image, while WavLM and BERT yield sequence-level embeddings from a series of an audio waveform and word tokens.

\section{Multimodal Sentiment Analysis by Modality-Specific Pre-Trained Encoders}
\label{sec:proposed}
\subsection{Model Structure}
\label{sec:proposed_model}
Let $\bm{S}_v, \bm{S}_a, \bm{S}_l$ be input data of visual, acoustic, and linguistic modalities, respectively, and $y$ be the corresponding sentiment value of the utterance.
Multimodal sentiment analysis is defined as the regression task of determining $y$ from $\bm{S}_v, \bm{S}_a, \bm{S}_l$,
\begin{align}
	\hat{y} = f\left(\bm{S}_v, \bm{S}_a, \bm{S}_l; \Theta \right),
\end{align}
where $\hat{y}$ is the predicted sentiment value, $f(\cdot)$ is the regression function determined by the regression model, and $\Theta$ is a parameter set of the regression model.

\begin{figure}[t]
	\centering
	\includegraphics[width=80mm]{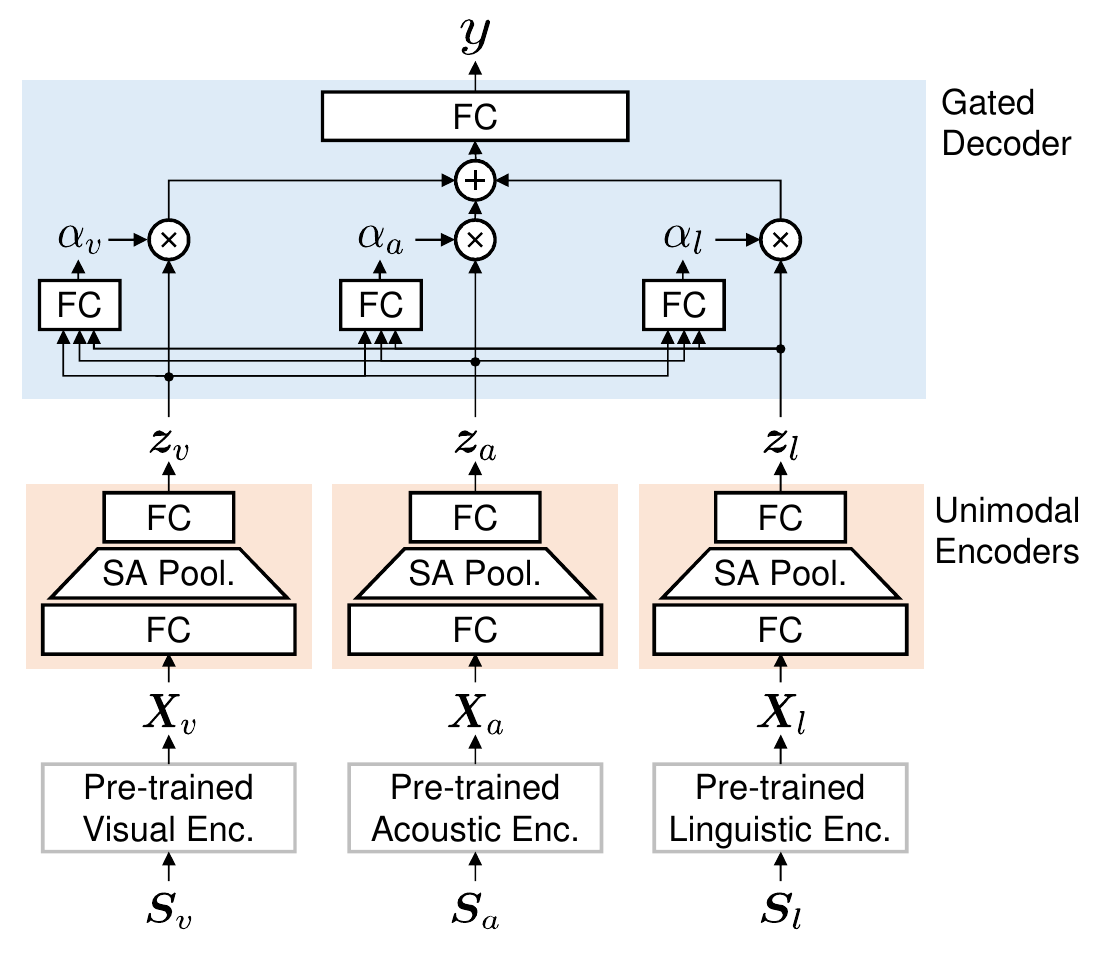}
	\vspace{-1em}
	\caption{Overview of the Unimodal Encoders and Gated Decoder~(UEGD) model.}
	\label{fig:proposed}
\end{figure}
This paper uses Unimodal Encoders and Gated Decoder~(UEGD) as the regression model, see Fig.~\ref{fig:proposed}.
The inputs of the UEGD model are modal-dependent low-level features $\bm{X}_v, \bm{X}_a, \bm{X}_l$ extracted from $\bm{S}_v, \bm{S}_a, \bm{S}_l$ with pre-trained encoders, see Section~\ref{sec:proposed_feat} for details.
UEGD extracts utterance-level embeddings of each modality $\bm{z}_v, \bm{z}_a, \bm{z}_l$, 
\begin{align}
	\bm{z}_m = \mathsf{UnimodalEncoder}\left( \bm{X}_m; \theta_m^{(e)} \right),
\end{align}
where $m \in \{v, a, l\}$ and $\mathsf{UnimodalEncoder(\cdot)}$ is a projection function from low-level features to the utterance-level embedding in each modality.
$\theta_m^{(e)}$ is a set of parameters of the unimodal encoder.
$\bm{z}_v, \bm{z}_a, \bm{z}_l \in \mathbb{R}^F$ and $F$ are dimensions of the embeddings.
The unimodal encoder consists of Fully-Connected~(FC) and Self-Attentive~(SA) pooling layers.
SA pooling layer allows hidden vectors having specific intervals in the utterance-level features to be strongly reflected, which is desirable in sentiment analyses since sentiment cues will appear in limited regions of input data.

The utterance-level embeddings are integrated and projected by the gated decoder to yield the predicted result, 
\begin{align}
	\hat{y} = \mathsf{GatedDecoder}\left(\bm{z}_v, \bm{z}_a, \bm{z}_l ; \theta^{(d)} \right), 
\end{align}
where $\mathsf{GatedDecoder}(\cdot)$ is a function projecting the estimated sentiment values from the embeddings.
$\theta^{(d)}$ is a set of parameters of the gated decoder.
The gated decoder has a gate layer that integrates the embeddings by the weighted sum, 
\begin{align}
	\alpha_m &= \sigma \left(\bm{W}_m \left[\bm{z}_v; \bm{z}_a; \bm{z}_l\right] + b_m \right), \\
	\bar{\bm{z}} &= \sum_{m\in\{v, a, l\}} \alpha_m \bm{z}_m, 
\end{align}
where $\alpha_m$ represents the gate weight of the specific modality $m$ calculated by the supervector concatenated by embeddings of all modalities.
$\{\bm{W}_m, b_m\} \in \theta^{(d)}$ is the set of weight and bias parameters.
$\sigma(\cdot)$ is a sigmoid function that constraints the gate weight to lie between 0 to 1.
The integrated embedding $\bar{\bm{z}}$ is input to the FC layers to obtain $\hat{y}$.
One of the advantages of the gated decoder is that it allows quantifying the contribution of each modality to the prediction result as the gate weight\footnote{In the preliminary experiments, the gate fusion performed the same as the concatenation of the utterance-level embeddings.}.

The model parameters $\Theta = \{\theta_v^{(e)}, \theta_a^{(e)}, \theta_l^{(e)}, \theta^{(d)}\}$ are optimized by L1 loss of ground truth $y$ and the predicted result $\hat{y}$.
Note that all the modality-specific pre-trained encoders are frozen during training since they have so many parameters that making them trainable would lead to overfitting.

Another advantage of the UEGD model is that it can be used for unimodal as well as multimodal cases on the same structure.
In the case of unimodal input, the utterance-level embeddings of the unused modalities $\bm{z}_m$ are set to $\boldsymbol{0} \in \mathbb{R}^F$.
Experiments have confirmed that the gate weights of the unused modalities approach zero, which means the UEGD model evaluates sentiment from specific unimodal inputs.

\subsection{Feature Extraction with the Pre-Trained Encoder}
\label{sec:proposed_feat}
\begin{figure}[t]
	\centering
	\includegraphics[width=50mm]{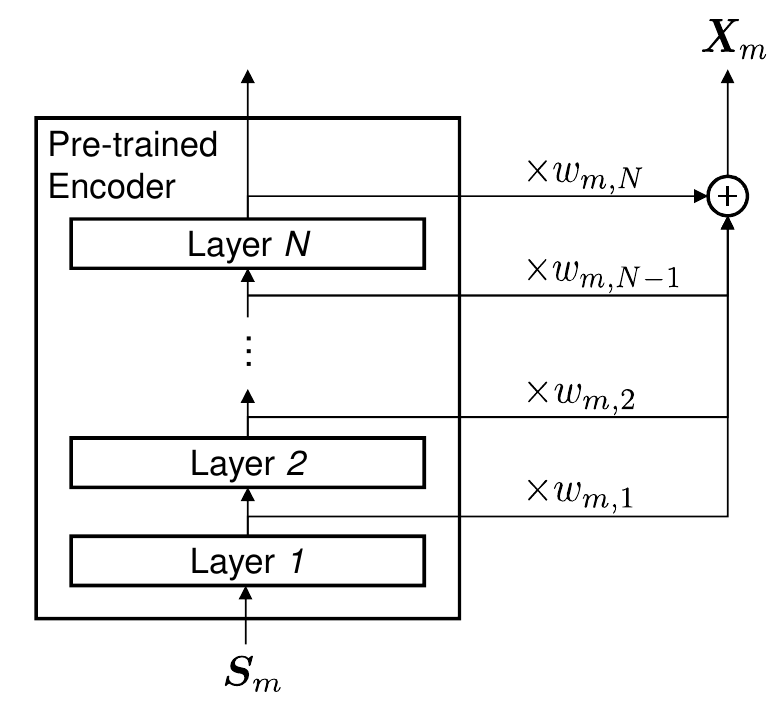}
	\vspace{-1em}
	\caption{The weighted sum of the pre-trained encoder outputs.}
	\label{fig:weighted_sum}
\end{figure}

Three types of the extracted features $\bm{X}_m$ yielded by each of the modality-specific pre-trained encoders are investigated;
the output, any intermediate output, and a weighted sum of the intermediate outputs.
The first two represent the output vectors of the final layer or the intermediate layers of the pre-trained encoder, respectively.
The weighted sum uses the weighted sum of the hidden states extracted from each layer of the pre-trained model as reported in the conventional work of~\cite{wavlm}, see Fig.~\ref{fig:weighted_sum}. 
The weights of the intermediate layers are parameters that are optimized simultaneously with the parameters of the UEGD model by the labeled data.

\begin{table}[t]
	\caption{The numbers of clips in the datasets.}
	\label{tbl:dataset}
	\vspace{-2em}
	\begin{center}
		\begin{tabular}{l|rrr|r} \hline
					    & \#~Train   & \#~Valid.  & \#~Test    & \#~Total   \\ \hline
			CMU-MOSI	& 1284      & 229       & 686       & 2199      \\
			CMU-MOSEI	& 16326     & 1871      & 4659      & 22856	    \\ \hline
		\end{tabular}
	\end{center}
\end{table}
\section{Experiments}
\label{sec:expr}
\subsection{Datasets}
\label{sec:expr_dataset}
Two public multimodal datasets were used in the experiments; CMU-MOSI~\cite{mosi} and CMU-MOSEI~\cite{CMUMOSEI}. 
Both contain short clips from YouTube videos with sentiment labels given by human annotators.
Each clip contains only one person and an utterance of a few seconds.
The label takes values from -3 to +3.
The total numbers of videos are 98 and 3178 on CMU-MOSI and CMU-MOSEI, respectively, which almost match the numbers of speakers.
The datasets were divided into training, validation, and test subsets as in the conventional studies
~\cite{MISA, SelfMM}.
The numbers of clips are shown in Table~\ref{tbl:dataset}.

\begin{table*}[t]
	\caption{Performance of the multimodal models. In the Feature column, Conv and Conv-BERT are sets of the conventional features including GloVe and BERT-Base features from linguistic modality, respectively. In Acc-2 and F-Score, the left of "/" is "negative/non-negative" score and the right is "negative/positive" performance. Results of $^\dagger$, $^\ddagger$ and $\diamond$ are from \cite{MISA}, \cite{Integrating_Multimodal_Information}, and \cite{SelfMM}, respectively.}
	\label{tbl:result_multimodal}
	\vspace{-2em}
	\begin{center}
		\begin{tabular}{ll|cccc|cccc} \hline
					    &               &  \multicolumn{4}{|c|}{CMU-MOSI}	                    & \multicolumn{4}{|c}{CMU-MOSEI}    \\ \cline{3-10}
					    & Feature       & MAE\,$\downarrow$       & Corr\,$\uparrow$      & Acc-2\,$\uparrow$         & F-Score\,$\uparrow$      & MAE\,$\downarrow$   & Corr\,$\uparrow$ & Acc-2\,$\uparrow$         & F-Score\,$\uparrow$ 	 	\\ \hline
			TFN~\cite{Tensor_Fusion_Network}$^\dagger$	    & Conv     & 0.970	    & 0.633	    & 73.9/-	    & 73.4/-    & -	        & -	        & -/-	& -/-	\\
			            & Conv-BERT     & 0.901	    & 0.698	    & -/80.8	    & -/80.7    & 0.593	& 0.700	& -/82.5	& -/82.1	\\
			MulT~\cite{MulT}$^\ddagger$	    & Conv          & 0.871	    & 0.698	    & -/83.0	    & -/82.8    & 0.580	& 0.703	& -/82.5	& -/82.3	\\
			            & Conv-BERT          & 0.861	    & 0.711	    & 81.5/84.1	    & 80.6/83.9    & -	& -	& -/83.5	& -/82.9 \\
			MISA~\cite{MISA}$^\diamond$	    & Conv-BERT     & 0.804	    & 0.764	    & 80.8/82.1	    & 80.8/82.0   & 0.568	& 0.724	& 82.6/84.2	& 82.7/84.0	\\
			MAG~\cite{Integrating_Multimodal_Information}$^\diamond$	& Conv-BERT     & 0.731	    & 0.789	    & 82.5/84.3     & 82.6/84.3   & 0.539	& 0.753 & \textbf{83.8}/85.2   & \textbf{83.7}/85.1	\\
			Self-MM~\cite{SelfMM}$^\diamond$		& Conv-BERT     & \textbf{0.713}     & \textbf{0.798}     & \textbf{84.0}/\textbf{86.0}   & \textbf{84.4}/\textbf{86.0}   & 0.530 & 0.765 & 82.8/85.2   & 82.5/85.3	\\ \hline
			UEGD        & Conv          & 0.953	    & 0.663	    & 76.3/77.4	& 76.3/77.4	& 0.598	& 0.683	& 78.9/81.3 & 79.2/81.0 \\
			            & Conv-BERT     & 0.886	    & 0.691	    & 78.6/79.9	& 78.5/79.9	& 0.543	& 0.748	& 81.2/84.6	& 81.7/84.5	\\ \cline{2-10}
			     		& Enc. output   & 0.850	    & 0.715	    & 79.4/80.8	& 79.3/80.8	& 0.519	& 0.776	& 82.5/85.8	& 82.8/85.7	\\
			     		& Enc. mid-best & 0.828	    & 0.748	    & 82.0/83.9	& 82.1/84.0	& \textbf{0.506}	& \textbf{0.790}	& 82.4/\textbf{86.1}	& 82.7/\textbf{86.0} \\
			 	    	& Enc. weighted & 0.818	    & 0.749	    & 80.4/82.3	& 80.4/82.3	& 0.510	& 0.785	& 82.3/85.7	& 82.7/85.6	\\ \hline
		\end{tabular}
	\end{center}
\end{table*}
\subsection{Setup}
\label{sec:expr_setup}
The pre-trained encoders of the individual modalities were as follows.
\textit{CLIP ViT-L/14}\footnote{https://github.com/openai/CLIP}, \textit{WavLM Large}\footnote{https://github.com/microsoft/unilm/tree/master/wavlm}, and \textit{BERT-Large-uncased}\footnote{https://pypi.org/project/pytorch-pretrained-bert/} were used in visual, acoustic, and linguistic modalities, respectively.
The training data consisted of 400\,M image-text pairs collected from the Internet for CLIP ViT, 94\,k hours of speech from audiobooks, podcasts and meetings for WavLM, 3.3\,B words from Wikipedia and books.
The numbers of hidden layers and embedding sizes were, for all encoders, 24 and 1024, respectively.
Face detection by MTCNN~\cite{Joint_Face_Detection} was applied to get a 256$\times$256-pixel face image before extracting visual embeddings.
For visual modality, the outputs of the \texttt{[class]} token position were used as the pre-trained encoder outputs for each frame.
Face detection and visual feature extraction were applied at 3~fps as in conventional visual feature extraction~\cite{CMUMOSEI, Tensor_Fusion_Network}.
The frame shift length of the audio features yielded by the encoders was originally 20~ms and subsampled to 1/10 to get 5~fps features.

Two sets of the conventional features were used: those the same as \cite{Multi_Attention_Recurrent}\footnote{https://github.com/A2Zadeh/CMU-MultimodalSDK} and \cite{SelfMM}\footnote{https://github.com/thuiar/MMSA/tree/master/src/MMSA}.
The former is composed of 35-dimensional FACET facial expression analysis results as visual, 74-dimensional COVEREP hand-crafted features as acoustic, and 300-dimensional GloVe embeddings as linguistic features.
The latter have the same visual and acoustic features and 768-dimensional BERT-Base output embeddings as linguistic features.
As in the previous work, only 20- and 5- dimensional features were used as the visual and acoustic features in CMU-MOSI.

The compared features from the domain-specific pre-trained encoders were the output~(Enc. output), the combinations of the single best of the intermediate outputs~(Enc. mid-best), and the weighted sum of the intermediate outputs~(Enc. weighted).
According to the correlation coefficient of the validation set in the unimodal scenario described in Section~\ref{sec:expr_result_unimodal}, we used the 15\,th, 21\,st, and 20\,th intermediate layers of visual, acoustic, and linguistic encoders respectively as Enc. mid-best in CMU-MOSI, while 19\,th, 21\,st, and 21\,st in CMU-MOSEI.
\begin{table*}[t]
	\caption{Performances with single modalities using the UEGD model. Bold means the best performances in each modality.}
	\label{tbl:result_indiv}
	\vspace{-2em}
	\begin{center}
		\begin{tabular}{ll|cccc|cccc} \hline
					    &               &  \multicolumn{4}{|c|}{CMU-MOSI}	            & \multicolumn{4}{|c}{CMU-MOSEI}    \\ \cline{3-10}
		Modality		& Feature       &  MAE\,$\downarrow$       & Corr\,$\uparrow$      & Acc-2\,$\uparrow$         & F-Score\,$\uparrow$      & MAE\,$\downarrow$   & Corr\,$\uparrow$ & Acc-2\,$\uparrow$         & F-Score\,$\uparrow$ \\ \hline
        Visual	        & Facet 	    & \textbf{1.431}	& \textbf{0.147}	& \textbf{54.0}/\textbf{52.9}	& \textbf{52.0}/\textbf{51.1}	& 0.802	& 0.278	& 68.3/66.1	& 66.1/62.4   \\ \cline{2-10}
    	                & Enc. output   & 1.468	& 0.033	& 47.3/45.5	& 41.1/39.5 & 0.771	& 0.403	& 69.6/70.6	& 69.5/69.4   \\
	                    & Enc. mid-best & 1.650	& -0.091	& 44.9/43.4	& 41.5/40.1 & 0.766	& 0.424	& 70.4/\textbf{72.2}	& \textbf{71.0}/\textbf{71.8}   \\
	                    & Enc. weighted & 1.630	& 0.074	& 51.2/50.2	& 49.7/48.8 & \textbf{0.762}	& \textbf{0.435}	& \textbf{70.7}/71.9	& \textbf{71.0}/71.3   \\ \hline
        Acoustic        & COVEREP       & 1.381	& 0.233	& 55.6/54.7	& 54.5/53.8 & 0.829	& 0.148	& 70.1/63.4	& 61.1/52.1   \\ \cline{2-10}
	                    & Enc. output   & 1.326	& 0.326	& 63.4/62.9	& 63.4/63.0	& 0.664	& 0.587	& 74.9/76.1	& 75.0/75.5   \\
	                    & Enc. mid-best & \textbf{1.098}	& \textbf{0.521}	& \textbf{70.0}/\textbf{70.9}	& \textbf{70.0}/\textbf{71.0}	& \textbf{0.605}	& \textbf{0.666}	& \textbf{77.4}/\textbf{80.4}	& \textbf{77.9}/\textbf{80.3}   \\
	                    & Enc. weighted & 1.247	& 0.406	& 65.9/66.0	& 65.8/66.0 & 0.635	& 0.629	& 77.0/78.7	& 77.1/78.1   \\ \hline
        Linguistic      & GloVe         & 0.963	& 0.634	& 76.6/77.7	& 76.6/77.7	& 0.616	& 0.661	& 78.4/80.1	& 78.6/79.6   \\
	                    & BERT-Base	    & 0.913	& 0.679	& 78.3/79.6	& 78.2/79.6	& 0.551	& 0.742	& 81.5/84.6	& 81.9/84.5   \\ \cline{2-10}
	                    & Enc. output   & 0.854	& 0.715	& 80.6/82.1	& 80.6/82.1	& 0.544	& 0.751	& 81.5/84.9	& 81.9/84.9   \\
	                    & Enc. mid-best & \textbf{0.821}	& \textbf{0.746}	& \textbf{81.8}/\textbf{83.6}	& \textbf{81.8}/\textbf{83.6}	& 0.540	& \textbf{0.760}	& \textbf{81.7}/\textbf{85.4}	& \textbf{82.2}/\textbf{85.3}   \\
	                    & Enc. weighted & 0.837	& 0.723	& 80.0/81.9	& 80.0/81.9 & \textbf{0.539}	& 0.758	& 81.1/84.9	& 81.6/84.9   \\ \hline
		\end{tabular}
	\end{center}
\end{table*}

The hyper-parameters of the proposed UEGD model were as follows.
The uni-modal encoder is composed of a fully-connected layer with 256 hidden units, self-attentive pooling with 4 heads, and a fully-connected layer with 128 hidden units, i.e., the size of the utterance-level embedding was 128.
The gated decoder consisted of a gating layer and two fully-connected layers with 128 and 1 hidden units, respectively.
We used the above hidden units in CMU-MOSEI and half of them in CMU-MOSI because the amount of the labeled training data was limited in CMU-MOSI.
Layer normalization and ReLU activation functions were applied after all fully-connected layers except for the output of the uni-modal encoder and the multimodal decoder.
The dropout rate was 0.2.
Batchsize was 16.
We used the Adam optimizer, and the learning rate was 0.0001 with warmup and cosine annealing.
Early stopping was applied via average loss of the validation set.
In the training step, masking was applied to a maximum of 20\% of the time and feature dimensions as in SpecAugment~\cite{SpecAugment} to prevent overfitting. 
We employed the same evaluation metrics as the conventional studies: Mean Absolute Error~(MAE), pearson correlation coefficient~(Corr), Accuracy~(Acc), and Weighted F1 score~(F-score).
The last two were the results of two-class classification tasks that predict negative/non-negative or negative/positive~(exclude zero).
In all experiments, we ran trials five times; the average performance is taken as the final result.
\subsection{Results}
\label{sec:expr_result}
\subsubsection{Evaluations of the Multimodal Model}
\label{sec:expr_result_multimodal}
Prediction performances by the conventional methods and the UEGD models with the conventional and pre-trained encoder output-based features are shown in Table~\ref{tbl:result_multimodal}.
Compared to the UEGD models, the encoder-based features showed better performances in MAE and Corr on both datasets.
Furthermore, all the encoder-based results with the UEGD models achieved better MAEs and correlation results than the conventional methods on CMU-MOSEI, even though they were based on a simple model and loss function.
On the other hand, the proposed method was inferior to several conventional methods on CMU-MOSI.
This indicates that the proposed UEGD model may not be optimized for small training data.
For example, MAG~\cite{Integrating_Multimodal_Information} is explicitly designed to use the linguistic features mainly for prediction, while the proposed method has to learn such knowledge from the training data, which is difficult if the training data is limited.
From these results, we consider that pre-trained encoders are more effective for multimodal sentiment analysis than the conventional features, especially in the case of a large amount of labeled training data.

\begin{figure}[t]
\vspace{-1.5em}
	\centering
	\subfloat[Conv-BERT]{\includegraphics[width=4.2cm]{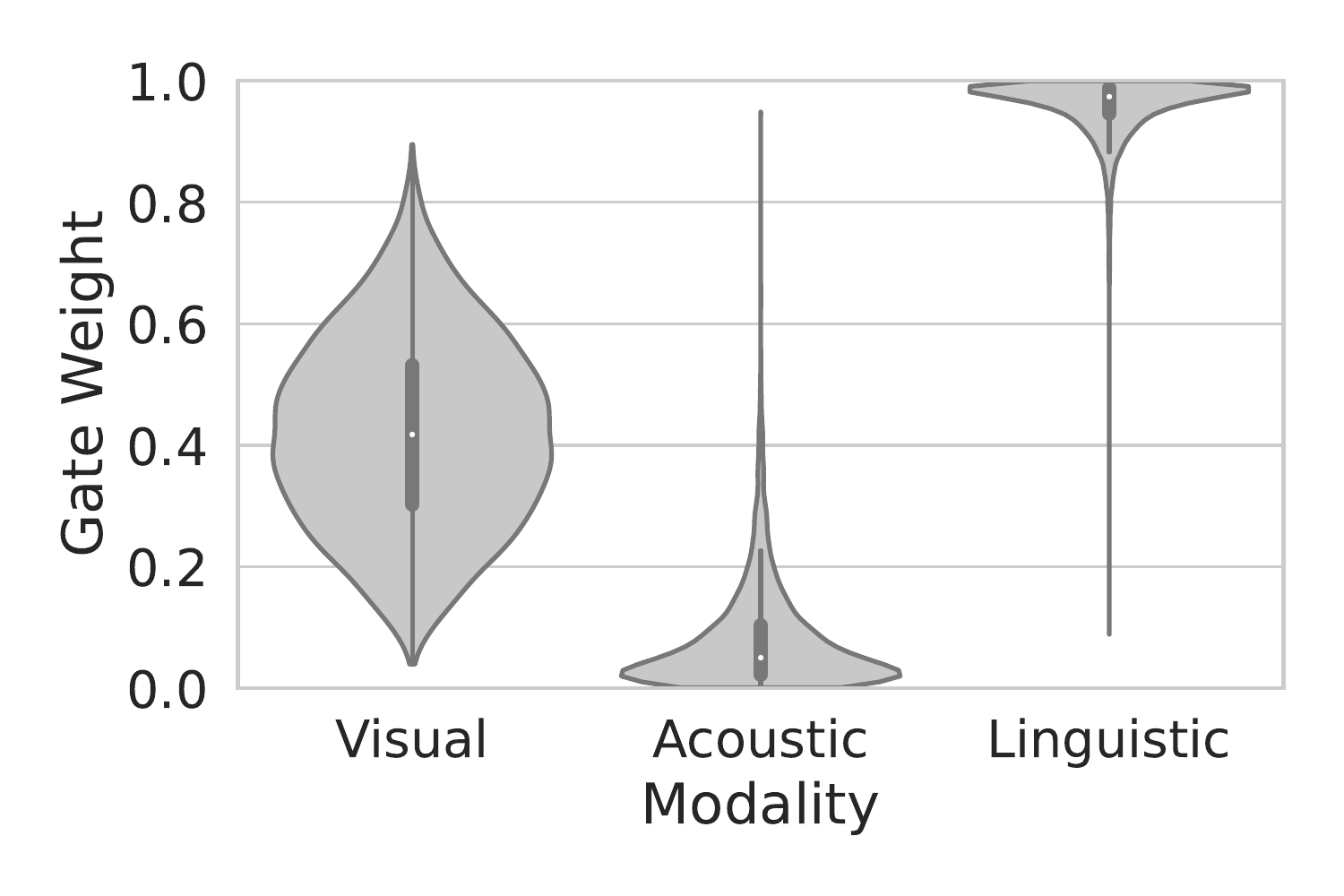}\label{fig:distrib_mmsa}}
	\subfloat[Enc. mid-best]{\includegraphics[width=4.2cm]{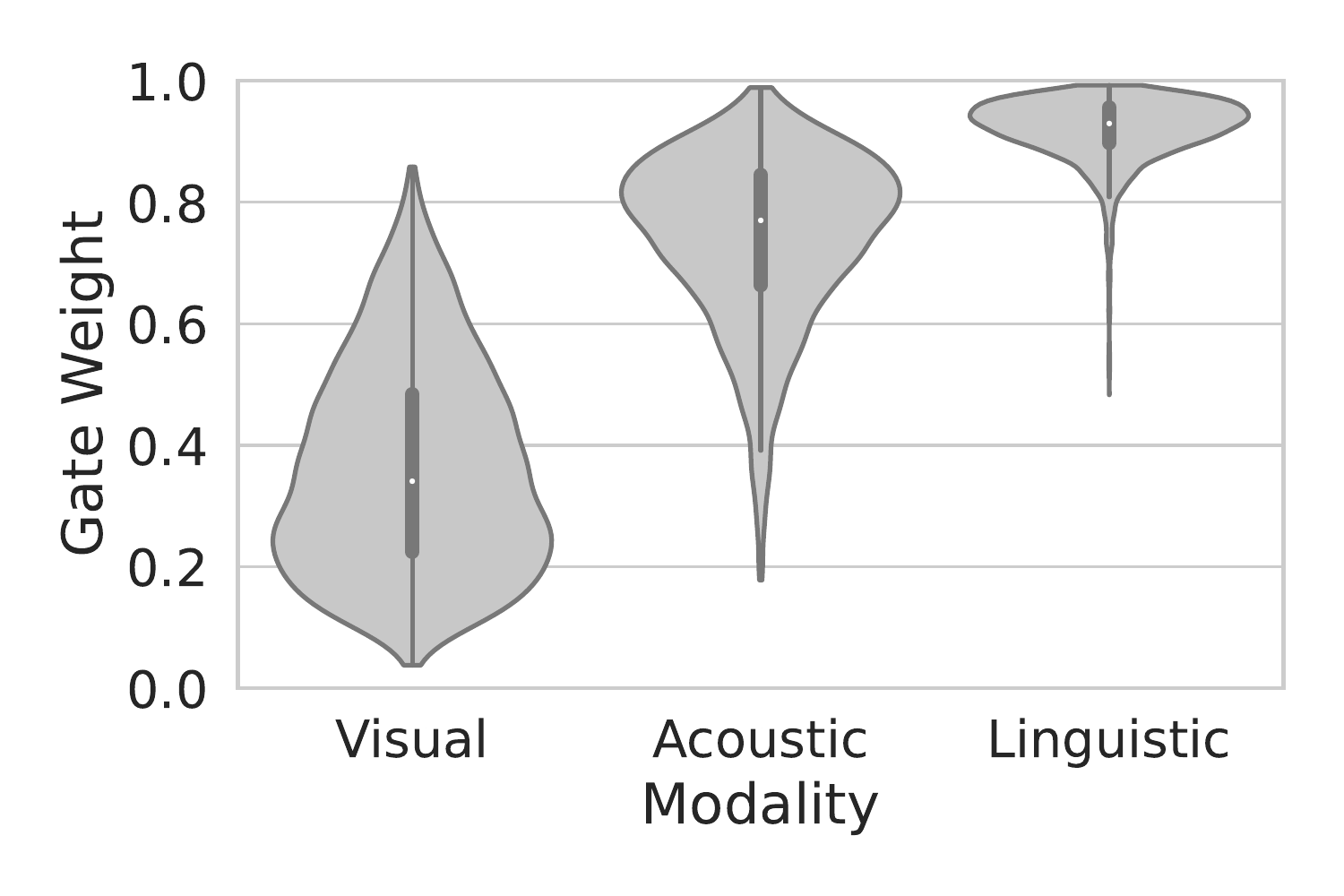}\label{fig:distrib_pretrain}}
	\vspace{-0.5em}
	\caption{Distributions of the gate weights of the UEGD models on the CMU-MOSEI dataset.}
	\label{fig:gate_distrib}
\vspace{-1em}
\end{figure}

The contributions of the individual modalities were also compared for both the conventional and the encoder-based features from the gate weights of the UEGD model.
The distributions of the gate weights in CMU-MOSEI are shown in Fig.~\ref{fig:gate_distrib}.
The gate weights of the acoustic modality using pre-trained outputs were distributed at higher values than those yielded by the conventional features.
This indicates that there was an improvement in the acoustic modality features, which results in better MSA performance.

\subsubsection{Evaluations of the Unimodal Model}
\label{sec:expr_result_unimodal}
The prediction results of individual modalities were also compared.
The results are shown in Table~\ref{tbl:result_indiv}.
Except for visual modality in CMU-MOSI, all the encoder-based features outperformed the conventional features in MAE and corr, especially acoustic modality in CMU-MOSEI.
With regard to the three encoder-based features, the output of the intermediate layers offers the best performance.
One possible reason for the lower performances of the weighted sum is the limited training data.
The amounts of training data for the downstream tasks in this work are smaller than those in the previous work~\cite{wavlm}, e.g., speech and speaker recognition, which may lead to overfitting of the weights of the intermediate outputs.

\begin{figure}[t]
	\centering
	\subfloat[video A]{\includegraphics[width=2.5cm]{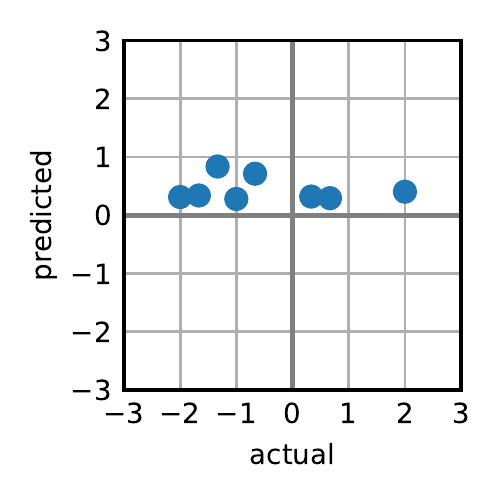}\label{fig:distrib_v_1}}
	\subfloat[video B]{\includegraphics[width=2.5cm]{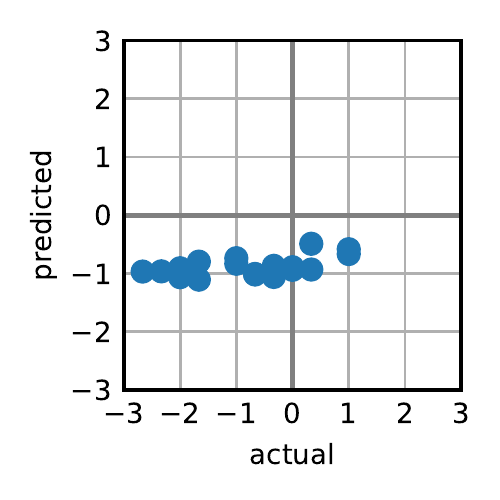}\label{fig:distrib_v_2}}
	\subfloat[video C]{\includegraphics[width=2.5cm]{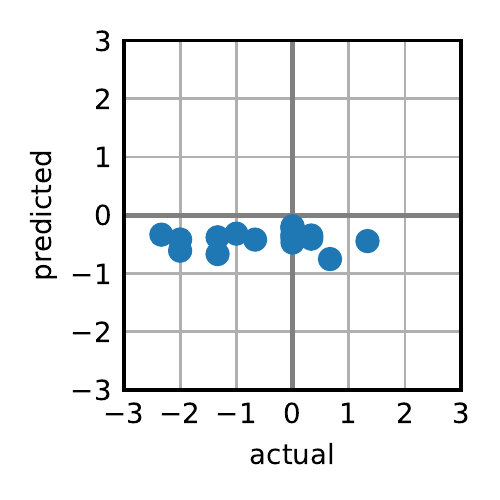}\label{fig:distrib_v_3}}
	\vspace{-0.5em}
	\caption{Prediction examples to the clips in the same video (speaker) by the pre-trained visual encoder.}
	\label{fig:eval_distrib_v}
	\vspace{-1em}
\end{figure}
\begin{figure}[t]
	\centering
	\subfloat[video A]{\includegraphics[width=2.5cm]{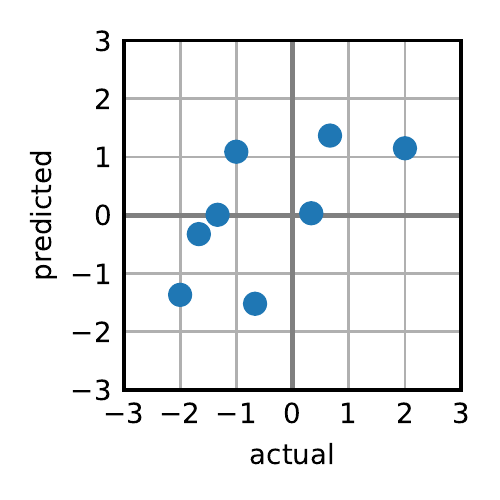}\label{fig:distrib_a_1}}
	\subfloat[video B]{\includegraphics[width=2.5cm]{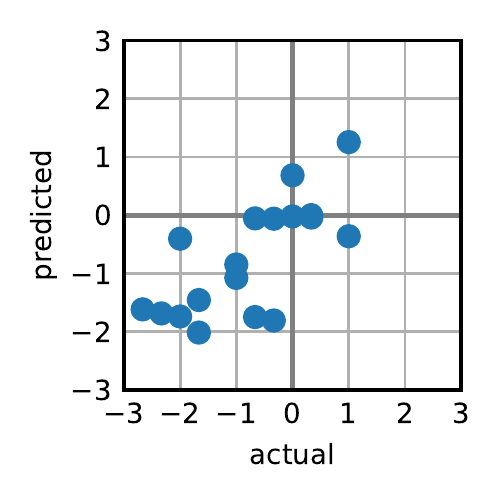}\label{fig:distrib_a_2}}
	\subfloat[video C]{\includegraphics[width=2.5cm]{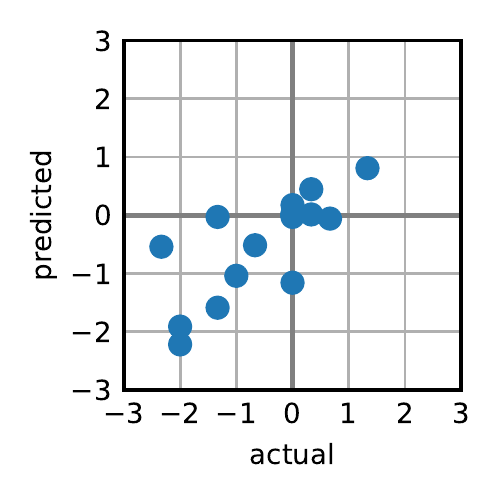}\label{fig:distrib_a_3}}
	\vspace{-0.5em}
	\caption{Prediction examples from the clips in the same video (speaker) by the pre-trained acoustic encoder.}
	\label{fig:eval_distrib_a}
\end{figure}

\begin{table}[t]
	\caption{Total and intra-video variances of the predicted values. The conventional linguistic feature was BERT-Base.}
	\label{tbl:result_var}
	\vspace{-2em}
	\begin{center}
		\begin{tabular}{ll|cc|cc} \hline
					&           &  \multicolumn{2}{|c|}{CMU-MOSI}  &  \multicolumn{2}{|c}{CMU-MOSEI}   \\ \cline{3-6}
				    & Modality          & Total     & Intra     & Total   & Intra       \\ \hline
			Conv.   & Visual	        & 0.661     & 0.430     & 0.060    & 0.014     \\
			        & Acoustic	        & 0.351     & 0.239     & 0.006    & 0.004     \\
			        & Linguistic       	& 2.009     & 1.606     & 0.737    & 0.427      \\ \hline
			Enc.   	& Visual            & 0.984     & 0.353     & 0.230    & 0.016      \\
			   	    & Acoustic          & 1.412     & 1.178     & 0.677    & 0.361      \\
			       	& Linguistic        & 2.005     & 1.612     & 0.750    & 0.414      \\ \hline
			\multicolumn{2}{l|}{\textit{Ground Truth}}  & 2.523     & 1.775    & 1.229   & 0.606      \\ \hline
		\end{tabular}
	\end{center}
	\vspace{-0.5em}
\end{table}
\begin{figure}[t]
 	\centering
 	\subfloat[CMU-MOSI]{\includegraphics[width=7.0cm]{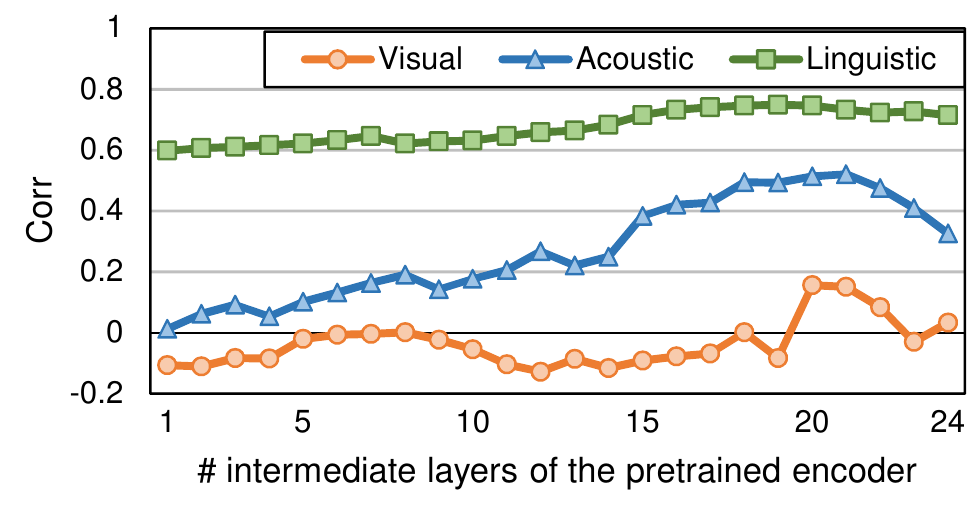}\label{fig:result_layers_mosi}}
\\ \vspace{-0.5em}
	\subfloat[CMU-MOSEI]{\includegraphics[width=7.0cm]{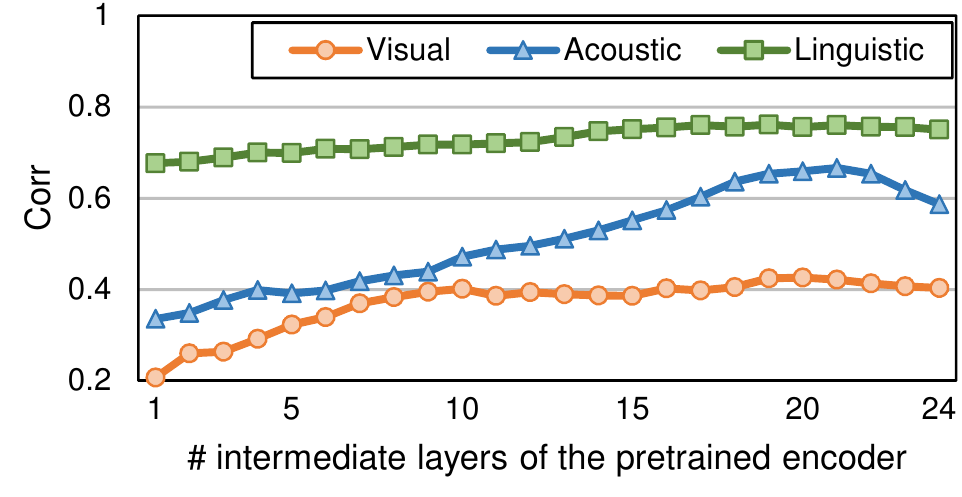}\label{fig:result_layers_mosei}}
	\vspace{-0.5em}
    \caption{Performances of the individual intermediate layers of the pre-trained encoders.}
	\label{fig:result_layers}
	\vspace{-1em}
\end{figure}
We conducted further analyses of the encoder output properties and found that the encoder outputs for visual modality may lead to extraction of speaker information rather than sentiment.
The prediction examples of the clips in the same videos, i.e., in the same speaker, by the pre-trained encoders in visual and acoustic modalities are shown in Fig.~\ref{fig:eval_distrib_v} and \ref{fig:eval_distrib_a}, respectively.
As shown in Fig.~\ref{fig:eval_distrib_v}, the model with a pre-trained visual encoder yielded similar prediction results for the same speaker, unlike the pre-trained acoustic encoder.
These characteristics appeared in the objective evaluations, total- and intra-video variances of the prediction results for each modality.
The results in Table~\ref{tbl:result_var} show that the visual encoder yielded much smaller intra-video variances than either the acoustic or linguistic encoder.
However, similar properties were observed in the conventional visual features.
Further investigation including evaluation of other datasets is required to elucidate the properties of the visual pre-trained encoder.

Finally, we discuss the characteristics of the intermediate layer outputs of the pre-trained encoders.
The correlation coefficients for each intermediate layer in the two datasets are shown in Fig.~\ref{fig:result_layers}.
The difference in performance of each intermediate layer was small for linguistic modality but significant for the other two modalities.
This may be because sentiment information appears as word (low-level) features in linguistic modality, while it appears as action units or speaking styles which are high-level features in visual and acoustic modalities.
For the acoustic modality, the highest accuracy was achieved when using around the 20th layer in both datasets.
It is considered that it is clearly better to use the middle layer of the second half of the pre-trained encoder than the output of the acoustic modality.
For the visual modality, the performances were almost flat after the  9\,th layer of the encoder on CMU-MOSEI.
This result suggests that there is no intermediate layer strongly associated with sentiment, at least for CLIP ViT-L.
Based on these results, best performance is likely to be achieved with the outputs of the second half of the domain-specific pre-trained encoders, especially for the acoustic modality.

\section{Conclusion}
\label{sec:conclu}
This paper investigated the effectiveness and implementation of domain-specific large-scale pre-trained encoders for MSA.
The regression model called UEGD was employed to compare the features in unimodal and multimodal scenarios.
Three types of features from large-scale pre-trained encoders were compared.
The findings from the experiments are as follows.
First, the large-scale pre-trained encoder yielded improved sentiment analysis performance in both unimodal and multimodal prediction models when a large amount of labeled training data was available.
Second, the pre-trained encoder is particularly effective for acoustic modality.
Third, the second half of any of the pre-trained encoders was most effective for MSA, rather than the output.
Future work includes further comparisons with other pre-trained encoders on other MSA datasets and the development of a multimodal decoder that can effectively utilize the pre-trained encoders.

\bibliographystyle{IEEEbibliostyle}
\bibliography{main}

\end{document}